\documentclass[pdflatex,sn-mathphys-num]{sn-jnl}

\usepackage{graphicx}%
\usepackage{multirow}%
\usepackage{amsmath,amssymb,amsfonts}%
\usepackage{amsthm}%
\usepackage{mathrsfs}%
\usepackage[title]{appendix}%
\usepackage{xcolor}%
\usepackage{textcomp}%
\usepackage{manyfoot}%
\usepackage{booktabs}%
\usepackage{algorithm}%
\usepackage{algorithmicx}%
\usepackage{algpseudocode}%
\usepackage{listings}%

\begin{document}

\title[Article Title]{DE-CGAN: Boosting rTMS Treatment Prediction with Diversity Enhancing Conditional Generative Adversarial Networks}


\author*[1]{\fnm{Matthew} \sur{Squires}}\email{matthew.squires@usq.edu.au}

\author[1]{\fnm{Xiaohui} \sur{Tao}}\email{xiaohui.tao@usq.edu.au}

\author[2]{\fnm{Soman} \sur{Elangovan}}\email{soman.elangovan@healthecare.com.au}

\author[3]{\fnm{Raj} \sur{Gururajan}}\email{Raj.Gururajan@usq.edu.au}

\author[4]{\fnm{Haoran} \sur{Xie}}\email{hrxie@ln.edu.hk}

\author[3]{\fnm{Xujuan} \sur{Zhou}}\email{xujuan.zhou@usq.edu.au}

\author[5]{\fnm{Yuefeng} \sur{Li}}\email{y2.li@qut.edu.au}

\author[1]{\fnm{U Rajendra} \sur{Acharya}}\email{rajendra.acharya@usq.edu.au}

\affil*[1]{\orgdiv{School of Mathematics, Physics and Computing}, \orgname{University of Southern Queensland}, \orgaddress{\city{Toowoomba}, \country{Australia}}}

\affil[2]{\orgname{Belmont Private Hospital}, \orgaddress{\city{Brisbane}, \country{Australia}}}

\affil[3]{\orgdiv{School of Business}, \orgname{University of Southern Queensland}, \orgaddress{\city{Springfield}, \country{Australia}}}

\affil[4]{\orgdiv{Department of Computing and Decision Sciences}, \orgname{Lingnan University}, \orgaddress{\city{Hong Kong SAR}, \country{China}}}

\affil[5]{\orgdiv{School of Computer Science}, \orgname{ Queensland University of Technology}, \orgaddress{\city{Brisbane}, \country{Australia}}}



\abstract{Repetitive Transcranial Magnetic Stimulation (rTMS) is a well-supported, evidence-based treatment for depression. However, patterns of response to this treatment are inconsistent. Emerging evidence suggests that artificial intelligence can predict rTMS treatment outcomes for most patients using fMRI connectivity features. While these models can reliably predict treatment outcomes for many patients for some underrepresented fMRI connectivity measures DNN models are unable to reliably predict treatment outcomes. As such we propose a novel method, Diversity Enhancing Conditional General Adversarial Network (DE-CGAN) for oversampling these underrepresented examples. DE-CGAN creates synthetic examples in difficult-to-classify regions by first identifying these data points and then creating conditioned synthetic examples to enhance data diversity. Through empirical experiments we show that a classification model trained using a diversity enhanced training set outperforms traditional data augmentation techniques and existing benchmark results. This work shows that increasing the diversity of a training dataset can improve classification model performance. Furthermore, this work provides evidence for the utility of synthetic patients providing larger more robust datasets for both AI researchers and psychiatrists to explore variable relationships.}


\maketitle

\section{Introduction}\label{sec:Introduction}
Depression is a highly prevalent and debilitating mental illness \cite{Schofield2019}. For some, treatment will lead to a reduction in their depression severity. However, approximately one-third of patients will see a minimal reduction in the severity of their depression \cite{Sforzini2021, Ionescu2015}. For many patients suffering from TRD, rTMS provides some relief. However, other patients will see little improvement in their depression severity. Given the variance in observed response rates, research is exploring the use of artificial intelligence (AI) techniques to predict treatment response to rTMS \cite{Squires2023, Adamson2022, Chen2022, Hopman2021, Bailey2019, Fan2019, Hasanzadeh2019, Zandvakili2019, Bailey2018, Koutsouleris2017, Drysdale2017, Erguzel2015}. These works aim to delineate between responders and non-responders to treatment before or early in the treatment cycle. Thus reducing the financial and psychological toll that ineffective treatments place on patients \cite{Hasanzadeh2019, Zandvakili2019}.

Personalising psychiatric treatments through the use of AI is a rapidly expanding area of research. Despite this interest, progress towards implementation has been slowed by challenges of model generalisation and access to appropriately sized data. \citet{Koppe2020} discuss some of the challenges of implementing deep learning (DL) in psychiatry. DL algorithms require large, diverse datasets to produce their best results, however, these large samples are generally not available in psychiatry \cite{Koppe2020}. When models are built on large, diverse and representative training sets the better the model will generalise to unobserved data \cite{Mumuni2022}. As such methods to address the lack of quality psychiatric data are required as a method to improve model generalisation. 

Enhancing fairness in AI is an important topic of research. Potential challenges to fairness of AI include data bias \cite{Ueda2023}. When limited examples of a phenomenon exists within the training set, classifiers trained on these datasets tend to generalize poorly \cite{Dunlap2023}. Furthermore,  the class imbalance problem is one of the most significant challenges in data mining \cite{Li2023}. Oversampling of the minority class in a dataset is one established strategy to balance datasets with few examples of the minority class \cite{Farahany2022, Turlapati2020}. However, these methods fail to address oversampling the feature space with distinct values which could belong to either class. Thus we seek a method which can both enhance the diversity of training dataset and create synthetic examples conditioned on class label. Conditional Generative Adversarial Networks (CGANs) offer an alternative to traditional generative networks by conditioning synthetic examples on additional information, such as class label. This paper proposes a novel methodology, Diversity Enhancing Conditional General Adversarial Network (DE-CGAN) for boosting the diversity of underrepresented samples conditioned on class labels. This proposed method seeks to boost the diversity of sparse regions of the feature space with synthetic examples belonging to either class, distinct from the problem of oversampling the minority class.

Recently, we showed a Deep Neural Network (DNN) could reliably predict rTMS treatment outcomes using fMRI connectivity features collected before treatment \cite{Squires2023}. However, this research also identified certain connectivity values, particularly values of functional connectivity between the Subgenual Anterior Cingulate Cortex and Occipital Cortex which a DNN consistently mislabeled. In an effort to address these misclassifications, we explore the use of generative networks to oversample examples of these difficult-to-classify patients and increase the diversity of the available data.  

Synthetic data is artificially created data \cite{Lucini2021}.  Broadly, two classes of methods exist for the generation of synthetic data: process-driven and data-driven \cite{Goncalves2020}. Process-driven methods "derive synthetic data from computational or mathematical models of an underlying physical process" \cite[p.2]{Goncalves2020}. However, process-driven methods cannot be implemented when a system cannot be modeled. As such, data-driven methods have surged in popularity in both research and public discourse. These data-driven methods use an underlying data distribution to create synthetic examples, which ideally, come from the same distribution \cite{Lucini2021, Chen2021}. The benefit of such methods is that generated data is assumed to contain the same characteristics as the original data but protects the privacy of the original data. This protection is especially important for data that contains sensitive information, such as medical data. However, as \citet{Jain2022} noted, that generated examples come from the same probability distribution is the best case. More likely, however, that generated examples are less diverse than the original data. As such, we explore methods for evaluating the quality of synthetic $(X,Y)$ pairs to increase the diversity of rTMS datasets to improve classification performance.
The current work aims to evaluate the use of synthetic data points to enhance the diversity of rTMS training data. We do this exploring the following research question.

\begin{enumerate}
    \item How does the use of synthetic rTMS patients impact classification model performance on a real test set?
    \item How does oversampling with synthetic patients of underrepresented fMRI connectivity features alter model performance on a held-out test set of real patients?
\end{enumerate}

In answering these questions the current paper makes the following contributions:
\begin{itemize}
    \item A novel framework we are calling Diversity Enhancing Conditional General Adversarial Network (DE-CGAN) for oversampling underrepresented fMRI connectivity features with synthetic values and their class labels.
    \item A diversity-enhanced rTMS dataset for the study of rTMS patterns of response to treatment. 
    \item Empirical experiments showing a diversity-enhanced training set improves model performance on a held-out test set of real examples.
\end{itemize}

This paper is structured as follows. Section \ref{sec:Related Work} provides a review of the current state-of-the-art techniques, for generating synthetic data. Including those used specifically for generating medical data. Section \ref{sec:Methods} the details of the constructed generative network are outlined. Section \ref{sec:ExperimentOverview} provides the experiment design and evaluation metrics with experiment results presented in Section \ref{sec:Results}. Finally, Section \ref{Discussion} provides a commentary on experiment results.

\section{Related Work}\label{sec:Related Work}
The class imbalance problem is one of the most significant challenges in data mining \cite{Li2023}. Class imbalance describes a situation where a target class is underrepresented in the training data compared to other classes \cite{Sharma2023,Fajardo2021}. When training sets lack diversity across classes performance can decline on the underrepresented examples \cite{Turlapati2020}. To address the class imbalance issue a variety of sampling and data augmentation techniques have been proposed \cite{Strelcenia2023}. 

Methods of data augmentation vary from traditional oversampling methods to more recent advances such as the use of AI to create synthetic examples. Synthetic Minority Oversampling TEchnique \cite[SMOTE]{Chawla2011} is one of the first oversampling methods to address the issue of class imbalance. SMOTE proposes using synthetic examples by oversampling the minority class. Synthetic examples are created by taking the distance between minority samples and creating a synthetic example in this region. However, some limitations of SMOTE have been identified, particularly when oversampling in regions surrounded by majority class samples which can introduce unnecessary noise \cite{Sowjanya2022,Yang2022}. 

Alternatively, generative networks and more recently, deep generative networks \citet[see]{Wang2022a}, have been used across a variety of sectors to generate synthetic examples. For example, in finance \cite{Sun2020}, health \cite{Arora2022}, and remote sensing \cite{Alkhalifah2022}. In healthcare, generative networks have largely been used to synthetically augment neuroimaging \cite{Arora2022} and electronic health records \cite{Li2023a}. For a detailed overview of GANs and their applications see \cite{Dash2023}. 

While SMOTE works to create linear combinations of existing minority samples modern AI techniques generate synthetic examples while capturing relationships between variables. Recently, \cite{Torfi2020} proposed a correlation-capturing Generative Adversarial Network (CorGAN) to synthesize electronic health records. Their method combines the use of a 1-dimensional Convolutional GAN with a convolutional autoencoder to discretize continuous values and produce the desired output. Convolutional autoencoders have been proposed as an alternative to the standard autoencoder \cite{Zhang2018}. In their work, \citet{Torfi2020} showed CorGan to outperform baseline models in creation of medical records and synthetic EEG data. Utilising these same generative techniques researchers have shifted from computer vision to the generation of synthetic medical information. For example, \citet{Choi2017} showed generative adversarial networks (GANs) were able to create realistic electronic health records. The architecture, named medGAN, produced predominantly realistic-looking electronic health records as judged by a doctor of medicine. More recently, \citet{Torfi2022} proposed convolution GANs for the generation of synthetic medical information. Their work utilised this architecture to construct synthetic medical data. \citet{Goncalves2020} explored several data-driven techniques for the generation of synthetic electronic health records. Their work evaluated the extent to which expected variable relationships were maintained using Surveillance, Epidemiology and End Results (SEER) a widely known cancer dataset. In contrast to \citet{Choi2017} and \citet{Torfi2022}, \citet{Goncalves2020} question the efficacy of GANs for creating realistic synthetic data. When comparing several techniques "the generative adversarial network-based model MC-MedGAN failed to generate data with similar statistical characteristics to the real dataset" \cite[p.30]{Goncalves2020}. These promising but contrasting findings motivate the exploration of GANs to model synthetic psychiatric data. 

Emerging evidence suggests data augmentation methods, such as generative networks, could be used to enhance the diversity of health datasets. For example, \citet{Behal2023} proposed Minority Class Rebalancing through Augmentation by Generative modeling (MCRAGE). The work proposes a conditional denoising diffusion probabilistic model to generate synthetic examples. MCRAGE aims to balance datasets of electronic health records based on demographic features such as race, gender and age. Their work shows a classifier trained on a synthetic dataset created by a diffusion improves classifier performance when compared to the original dataset.

DL is positioned to significantly disrupt healthcare. However, progress towards the implementation of AI has been slowed by the limited availability of diverse datasets. One possible strategy to overcome these barriers is the use of synthetic data \cite{Chen2021}. Synthetic data may have the potential to create more diverse datasets \cite{Savage2023}. To date the use of GANs in healthcare has focused largely on the generation of synthetic images \cite{Arora2022}. This motivates our work to explore the use of generative techniques to augment psychiatric data. Existing solutions address imbalances in the class label $y$. However, these methods fail to target under represented regions in feature spaces $[x_1, x_2, x_3, \cdots, x_n]$. \citet{Behal2023} provides one example of balancing imbalanced data based using feature space variables. This motivates our work to balance feature space variables in psychiatry and evaluate the impact this has on a classifier trained on a synthetic dataset

\section{Methods}\label{sec:Methods}
\subsection{Problem Statement}
Existing research has shown fMRI connectivity features can reliably predict the outcome of rTMS treatment using a DNN \cite{Squires2023}. As part of this existing work we identified repeatedly mislabelled examples share common characteristics. To address this issue, the current work proposes using synthetically generated patients to enhance the diversity of a training set in these difficult to classify regions. Previous work has explored algorithmic methods to oversample minority classes in imbalanced datasets. In contrast, this work seeks to generate under represented examples and their class labels to improve model performance. The current work presents DE-CGAN a novel method for enhancing the diversity of rTMS training sets. DE-CGAN seeks to balance datasets by oversampling difficult to classify regions of connectivity between the Subgenual Anterior Cingulate Cortex and Occipital Cortex.

\subsection{Problem Definition}
The current work aims to show our proposed framework DE-CGAN can boost the diversity of our rTMS patient dataset with synthetic patients. Through enhancing the diversity of the training set we seek to improve the performance of a classification model which predicts rTMS patient outcomes using fMRI functional connectivity measures. 

To test this empirically we evaluate the impact of augmenting our training data with varying proportions of synthetic patients. We use this hybrid dataset to train a new classifier and test its performance, and hence the quality of our synthetically generated examples on a test set of real examples.

For our experiments, we define real patients and their associated fMRI connectivity features as follows: \[D_{real} = \{(x_i, y_i)\}_{i=1}^N\] and \[x_i = [FP, OC, SPL, COC, lOCL, lOCR]\]

DE-CGAN generates synthetic patients and their class labels: \[D_{synth} = \{(x'_j, y'_j)\}_{j=1}^M\] and \[x'_j = [FP', OC', SPL', COC', lOCL', lOCR']\]

Combing the datasets let the resultant dataset be defined as:

\[D_{hybrid} = D_{real} \cup D_{synth}\]

Where $\alpha$ defines the proportion of synthetic patients within the training data set, such that:

  \[D_{hybrid}(\alpha) = \frac{D_{synth}}{ D_{real} } \]

\begin{table*}
    \centering
    \caption{Symbol Descriptions}
    \begin{tabular}{c|p{8cm}}
    \hline
        Symbol & Description  \\
        \hline
        $D_{real}$ & Original Dataset from \cite{Hopman2021} \\
        \hline
        $D_{synth}$ & Synthetic examples created by DE-CGAN\\ 
        \hline
        $D_{hybrid}$ & Training dataset denoted by $D_{real} \cup D_{synth}$ \\ 
        \hline
        $(x,y)$ & An original patient's features with class label\\
        \hline
        $(x', y')$ & Synthetic patient and class label\\
        \hline
        $\alpha$ & DE-CGAN hyperparamater, sets the proportionality of$D_{synth}$ to $D_{real}$\\
        \hline
        FP & Functional connectivity measure between Subgenual Anterior Cingulate Cortex and Frontal Pole\\
        \hline
        OC & Functional connectivity measure between Subgenual Anterior Cingulate Cortex and Occipital Cortex \\
        \hline
        SPL & Functional connectivity measure between Subgenual Anterior Cingulate Cortex and Superior Parietal Lobule \\
        \hline
        COC & Functional connectivity measure between Subgenual Anterior Cingulate Cortex and Centeral Opercular Cortex\\
        \hline
        lOCL & Functional connectivity measure between Subgenual Anterior Cingulate Cortex and Left Lateral Occipital Cortex\\
        \hline
        lOCR & Functional connectivity measure between Subgenual Anterior Cingulate Cortex and Right Lateral Occipital Cortex\\
        \hline

    \end{tabular}
    
    \label{tab:Notation}
\end{table*}

\subsection{Conceptual Model}
This section describes the conceptual model of the framework deployed to address the problem described above. Based on the results of our literature review this is the first model deployed to generate synthetic rTMS patients with the aim of enhancing the diversity of our dataset. Through this oversampling we aim to improve predictive model performance to improve the generalisability of our model to unseen data. 

\begin{figure*}[ht]
     \centering
    \fbox{\includegraphics[scale = 0.65]{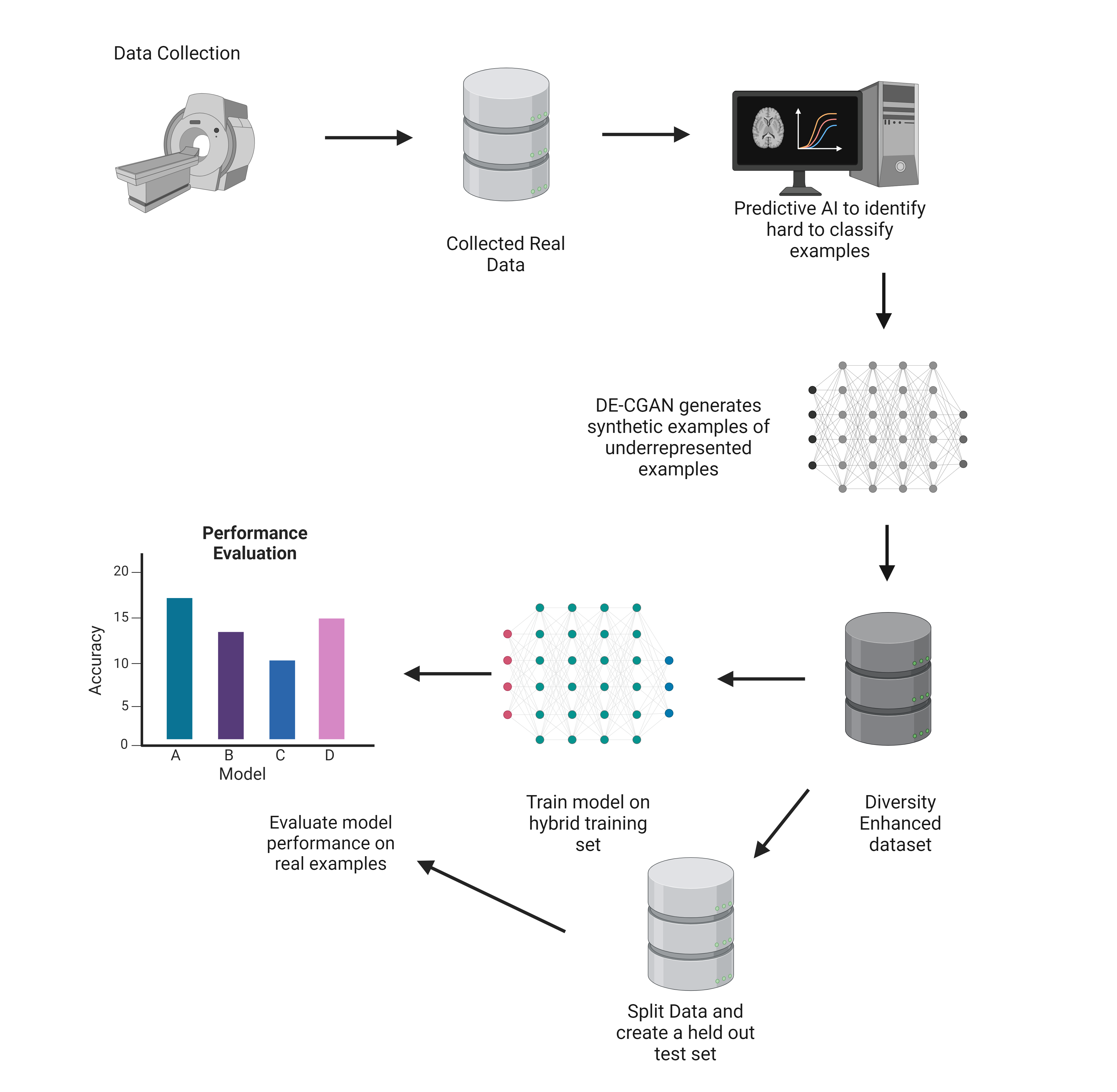}}
    \caption{DE-CGAN conceptual model (Created with BioRender.com)}
    \label{fig:Conceptual Model}
\end{figure*}

Figure \ref{fig:Conceptual Model} describes the conceptual framework that underpins our work. Previously. in \citet{Squires2023} we showed that a DNN trained on fMRI connectivity features could reliable predict rTMS treatment outcomes before treatment begins. As part of this work we identified a portion of the dataset which was regularly mislabelled. Using these results, we use these mislabeled examples to train a conditional general adversarial network (CGAN) to oversample the mislabelled examples with class labels. These works are distinct to existing works which oversample the minority class. By augmenting and extending the dataset with synthetic patients each with their own class label.

To evaluate our models performance we leverage our previously described results to evaluate the impacts of including synthetic examples in model training data. The details of which are described in Section \ref{sec:EvaluationMetrics}.

\subsection{Data Augmentation}
Data Augmentation is a class of regularization techniques aimed at improving model performance by making modifications to the training data \cite{Mumuni2022}. One data augmentation technique involves the use of generative models to generate synthetic examples within from the same distribution as the training data \cite{Koppe2020}. \citet{Mumuni2022} refers to these as 'data synthesis methods' methods which create new synthetic training examples generally through methods such as GANs or VAEs. 

To date data augmentation research has largely focused on image \cite{Naveed2021} and signal-based tasks \cite{Lashgari2020}. In the medical space these data augmentation techniques have largely been applied to medical imaging \cite{Garcea2023, Goceri2023}. For example, \citet{FridAdar2018} showed that synthetically augmenting a dataset of computed tomography (CT) images improves model performance. While to date, the majority of research has focused on health-related data augmentation, however, to our knowledge, little work has explored data augmentation in psychiatry.

\subsubsection{GAN}
The classical GAN was first proposed by \citet{Goodfellow2014}, which harnesses game theory to generate synthetic examples based on the training data set. The game takes place between two networks, the generator, and the discriminator. In classical architecture, the role of the generator is to create realistic examples derived from the training data set. The goal of the generator is to deceive the discriminator into mislabeling fake examples as real. In contrast, the goal of the discriminator is to accurately discern fake examples. Through training, the game reaches a state of Nash equilibrium where neither the discriminator nor the generator is able to further advance their position. Nash \cite{Nash1951} in the seminal work showed all 2-person zero-sum games have such an equilibrium point. In theory, the generated examples should mirror the probability distribution of the training data on which the network was trained. However, many in the research community question this assertion \cite{Jain2022,Arora2018, Arora2017}. The first generative network was described by \citet{Goodfellow2014}. The theoretical proof provided by \citet{Goodfellow2014} assumes a non-parameterized network of infinite capacity. However, in practice network size cannot be infinite. This bounding of network size has led some \cite{Jain2022, Arora2018, Arora2017} to question whether the generated data is representative of the data on which the model was trained. \citet{Arora2018} asserts even for popular GAN variants mode collapse remains a significant issue. 
The manifestation of the reduction in the diversity of generated examples has significant implications. Recently, \citet{Jain2022} demonstrated the inherent distributional decay and biases of several popular GAN variants. In novel works, \citet{Jain2022} used generative networks to create images of engineering staff members. Generated images were then assessed by human raters along with race and gender. \citet{Jain2022} reports synthetic images of engineering professors were more likely to have lighter skin tones and masculine features when compared to the original distribution of actual images. These works add tangible examples of differences between synthetic data and the original data distribution.

\subsubsection{CGAN}
Extending the original GAN the CGAN was first proposed by \cite{Mirza2014}. The CGAN involves training the original GAN on some additional information, such as a class label \cite{Mirza2014}.  The strength of the CGAN over the original GAN is by conditioning on class labels outputs will also have class labels. For example, \citet{Sun2023} show a novel cGAN architecture to generate labeled facial expression images. \citet{Mert2022} showed the use of synthetic datasets developed using CGANs improved classifier performance on several medical datasets. Compared to other data augmentation frameworks the CGAN is prefered for this problem as it allows for the conditioning synthetic examples on class labels. Thus capturing important relationships between feature variables and class label.

\subsection{DE-CGAN Model Architecture and Hyperparamaters}
This Section provides a detailed overview of our proposed DE-CGANs model architecture. Algorithm \ref{alg:DECGAN} describes the workflow for our DE-CGAN framework. As part of this method we identify frequently mislabelled examples when evaluating a DNN trained and evaluated on real examples. We then use these frequently mislabelled examples as training data for a CGAN. This produces synthetic examples in difficult to classify regions of the training dataset. The output of our DE-CGAN is a diversity enhanced training dataset for evaluation.

\begin{algorithm}
\caption{Diversity Enhancing CGAN Workflow}
\label{alg:DECGAN}
\begin{algorithmic}[1] 

\Statex \textbf{Input:} $D_{\text{real}}$
\Statex \textbf{Output:} $D_{hybrid}$

\Statex \textbf{Parameters:}  $\alpha$, DNN, CGAN, 

\State Train Deep Neural Network with $D_{\text{real}}$
\For{each input $x$ in $D_{\text{real}}$}
    \State Predict label $y$ using the Deep Neural Network
\EndFor
\State Pass the labeled data to \textbf{Mislabelled}
\Function{Mislabelled} {$x, y$}
\State Identify Mislabelled cases
\State Initialize Conditional GAN with Mislabelled examples
\EndFunction
\State \textbf{Conditional GAN Details:}
\State \quad \textbf{Generator \( G \):} Receives a noise vector \( z \) and label \( y \), generates synthetic data samples \( G(z, y) = x^* \)
\State \quad \textbf{Discriminator \( D \):} Receives both real data samples $(x|y, y)$ and synthetic data samples $(x^*|y, y)$, outputs a probability (via sigmoid) indicating the likelihood of the sample being real
\State Train Conditional GAN (Generator \( G \) and Discriminator \( D \)) iteratively
\For{a specified number of iterations or until convergence}
    \State \quad Train Discriminator \( D \) to distinguish between real and generated data
    \State \quad Train Generator \( G \) to create synthetic data that fools Discriminator \( D \)
\EndFor
\State Generate $\alpha$ synthetic cases using CGAN Generator $D_{synth}$
\State Split $D_{\text{real}}$ into a training set $D_{\text{train}}$ and a test set $D_{\text{test}}$
\State Transform $D_{train} \cup D_{synth}$ = $D_{hybrid}$

\end{algorithmic}
\end{algorithm}




The hyperparameters of both the generator and discriminator are shown in Table \ref{tab:Gen_hyperparamaters} and Table \ref{tab:Dis_hyperparamaters} respectively. The architecture of these networks is motivated by the architecture in \cite{Strelcenia2023} who adopted similar hyperparameters in designing a CGAN to generate synthtetic breast cancer data

\begin{table}
    \centering
     \caption{Generator Hyperparameters}
    \label{tab:Gen_hyperparamaters}
    \begin{tabular}{c|c}
    \hline
        Hidden Layers & 2  \\
        \hline
        Layer Width &  128, 64 \\
        \hline
        Activation Function & Leaky reLU \\
        \hline
        Latent Dimensions & 100\\
        \hline
      
    \end{tabular}
   
\end{table}

\begin{table}
    \centering
     \caption{Discriminator Hyperparameters}
    \label{tab:Dis_hyperparamaters}
    \begin{tabular}{c|c}
    \hline
        Hidden Layers & 2  \\
        \hline
        Layer Width &  64, 32 \\
        \hline
        Activation Function & Leaky reLU \\
        \hline
        Loss Function & Binary Cross-entopy \\
        \hline
       Optimizer & Adam\\
        \hline
        Learning Rate & 0.0002 \\
        \hline
    \end{tabular}
   
\end{table}

\section{Experiment Overview}\label{sec:ExperimentOverview}
\subsection{Experiment Design}
This section outlines the experiments used to evaluate the quality of datasets augmented by our DE-CGAN. The current work aims to show that augmenting training data with synthetically generated underrepresented examples improves the performance of a binary classification model. This work seeks to show that augmented training data is superior to the original training data. 

As part of our experiments, we evaluate the performance of varying proportions of augmented data on classifier performance. These experiments compare the performance of a classifier trained using data augmented by a traditional GAN, our DE-CGAN and the original dataset. We test these training datasets empirically using 200 runs of Monte Carlo simulation. During each simulation, a DNN is trained using an augmented dataset with a validation set of 20\% used during each training iteration. Early stopping is used to terminate the training process once model performance stabilises. The performance of each model is tested on a held-out test set of real examples. Further details of this methodology are described in Section \ref{sec:EvaluationMetrics}.

\subsection{Dataset}

The original data used in this work is publicly available in \citet{Hopman2021, Hopman2021a}. The data includes several fMRI features, along with a patient's rTMS treatment outcomes. Further detailed summary statistics of this dataset, including associated ethics approval, can be found in \cite{Hopman2021a}. The input variables used from this data are detailed in \ref{tab:Notation}.

\subsection{Evaluation metrics}\label{sec:EvaluationMetrics}

This section further outlines the methods and metrics used to evaluate the quality of augmented training sets:

\begin{itemize}
    \item \textbf{Classification} Evaluating generative networks is an open problem \cite{Borji2022}. \citet{Esteban2017} proposed a novel framework for evaluating the quality of synthetic examples. Train Synthetic Test Real (TSTR) involves training a classifier on synthetic examples and evaluating its performance on real examples. If the distribution of the synthetic data matches the real data then we would expect the performance of a classifier trained using synthetic examples to perform similarly to that of a classifier trained on the real dataset. This method has several benefits over training on real sets and evaluating on synthetic data. Given the limitations of many generative methods is mode collapse, that is, when synthetic examples become less diverse than the original training data. Hence, the performance of a model on trained on real data and tested on synthetic data may overestimate the quality of these synthetic examples. 

    \item \textbf{Hypothesis Testing} We propose the use of two-tail proportion test to evaluate any differences between the proportion of optimal solutions found using training sets augmented with varied proportions of synthetic data. 

\end{itemize}

\subsubsection{Classification Model and Evaluation}
Using the TSTR framework evaluation metrics are required to test the classifier performance on the held-out test set. Given the aim of this work is to evaluate the impact of augmenting training data with synthetic examples it is important the classifier remains constant across examples. The model hyperparameters are shown in Tab \ref{tab:model_hyperparamaters}. In each simulation the network is retrained after shuffling the augmented data set.

\begin{table}
    \centering
     \caption{Model Hyperparameters}
    \label{tab:model_hyperparamaters}
    \begin{tabular}{c|c}
    \hline
        Hidden Layers & 4  \\
        \hline
        Layer Width &  10 \\
        \hline
        Activation Function & reLU \\
        \hline
        Loss Function & Binary Crossentopy \\
        \hline
        Regularisation Layers & 4 \\
        \hline
        Test set size & 20\% \\
        \hline
        Epochs & 2000 or until early stopping criteria met \\
        \hline
    \end{tabular}
   
\end{table}

To evaluate performance we use commonly used deep learning metrics: accuracy, described in Equation \ref{eq:Acc}, balanced accuracy, described in Equation \ref{eq:BA} and f1-score in Equation \ref{eq:f1}. Both balanced accuracy and f1 score are selected as they consider classification performance across classes.

\begin{equation}\label{eq:Acc}
\text{Accuracy} = \frac{\text{TP} + \text{TN}}{\text{N}}
\end{equation}

\begin{equation}\label{eq:BA}
\text{Balanced Accuracy} = \frac{1}{2} \left( \frac{\text{TP}}{\text{TP} + \text{FN}} + \frac{\text{TN}}{\text{TN} + \text{FP}} \right)
\end{equation}

\begin{equation}\label{eq:f1}
F_1 = 2 \times \frac{\text{Precision} \times \text{Recall}}{\text{Precision} + \text{Recall}}
\end{equation}

\subsubsection{Hypothesis Testing}
Further to the classification metrics described above we use hypothesis testing to compare the amount of optimal solutions found using varying proportions of augmented data. The proportion test, described in Equation \ref{equation:Proportion}, can evaluate the extent to which differences in proportions between the amount optimal solutions are found

\begin{equation}
\label{equation:Proportion}
z = \frac{(P_{gan}-P_{dist})^2}{SD(P_{gan})}
\end{equation}
Where
\[SD(P_{gan}) = \sqrt{\frac{P_{gan}\cdot{Q_{gan}}}{n}}\]

\section{Results}\label{sec:Results}

This section describes the results of our empirical experiments to evaluate the effectiveness of our proposed model DE-CGAN, and the synthetic patients generated by our model.

As described above the performance of our proposed method was evaluated using 200 monte carlo simulation. The average performance and standard deviation of these models is described in Table \ref{tab:results}. These results show a hybrid dataset of 10\% synthetic patients from DE-CGAN to be the best performing on a held out test set. Followed by a hybrid dataset of 5\% synthetic patients. These models including synthetic examples created by DE-CGAN Followed by our benchmark model presented in \citet{Squires2023} with the original dataset described in \cite{Hopman2021} and no synthetic examples.

\begin{table*}[h]
\centering
\caption{Mean model performance metrics and standard deviations}
\label{tab:results}
\begin{tabular}{@{}llll@{}} 
\toprule
Model           & Accuracy & F1-Score & Balanced Accuracy \\ 
\midrule
DE-CGAN $(\alpha=0.05)$  & 0.9280 (0.0664) & 0.9310 (0.6666) & 0.9286 (0.0661)  \\
DE-CGAN $(\alpha=0.10)$  & \textbf{0.9360 (0.0662)} & \textbf{0.9414 (0.0589)} & \textbf{0.9355 (0.0684)}  \\
DE-CGAN $(\alpha=0.15)$  & 0.9073 (0.0817) & 0.9085 (0.0908) & 0.9085 (0.0802)  \\
DE-CGAN $(\alpha=0.20)$  & 0.9112 (0.0787) & 0.9124 (0.0806) & 0.9130 (0.0775)  \\
\citet{Squires2023}      & 0.9227 (0.0696) & 0.9276 (0.0647) & 0.9224 (0.0708)  \\
CGAN $(\alpha=0.05)$     & 0.9246 (0.0617) & 0.9280 (0.0610) & 0.9248 (0.0613)  \\
CGAN $(\alpha=0.10)$     & 0.9134 (0.0767) & 0.9189 (0.0735) & 0.9128 (0.0773)  \\
CGAN $(\alpha=0.15)$     & 0.9023 (0.0858) & 0.9100 (0.0813) & 0.9007 (0.0874)  \\

\bottomrule
\end{tabular}
\end{table*}

\begin{figure*}[h]
    \centering
    \includegraphics[scale=0.5]{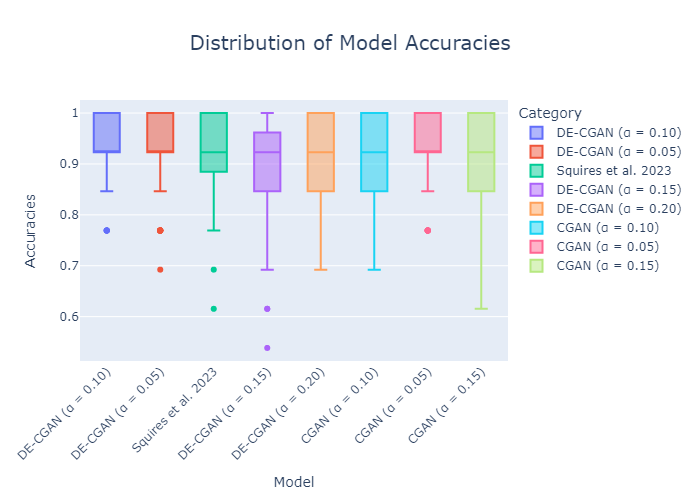}
    \caption{Box plot showing the distribution of accuracies obtained using various algorithms.}
    \label{fig:resultsBoxPlot}
\end{figure*}

Visually the distribution of accuracies is displayed in Figure \ref{fig:resultsBoxPlot}. This figure visually shows changes in accuracies across models through the empirical experiments with outliers. Visually we see similarities between the DE-CGAN $(\alpha=0.05)$ and DE-CGAN $(\alpha=0.10)$ distributions with our benchmark model \citet{Squires2023} and the baseline model CGAN $(\alpha=0.05)$.

\begin{figure*}[h]
    \centering
    \includegraphics[scale =0.5]{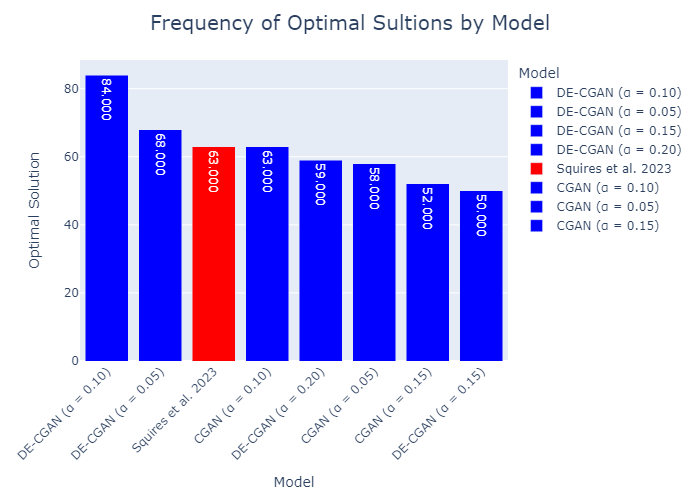}
    \caption{Bar chart showing the frequency of optimal solutions by algorithm.}
    \label{fig:resultsBarChart}
\end{figure*}

In further analysis of our results, we report the frequency of optimal solutions obtained by each algorithm. These results are shown in Figure \ref{fig:resultsBarChart}. From these results, we see both DE-CGAN $(\alpha=0.05)$ and DE-CGAN $(\alpha=0.10)$ are the only models to obtain the optimal solution more frequently than the benchmark model in \citet{Squires2023}. 

To formally compare the proportion of optimal solutions found by each model we use the proportion test to compare the frequencies against our benchmark model. From Table \ref{tab:proportion_test} we see the only model which varies significantly is DE-CGAN $(\alpha=0.10)$ which shows using our DE-CGAN to create a training set made up of 10\% synthetic cases produces performance above that from using the original dataset without synthetic data. 

\begin{table}[h]
\centering
\caption{Proportion Test Results}
\label{tab:proportion_test}
\begin{tabular}{@{}llll@{}} 
\toprule
Model Name       & Z      & P       & Significant \\ 
\midrule
DE-CGAN $(\alpha=0.05)$  & -0.5327 & 0.59612 & $p>0.05$\\
DE-CGAN $(\alpha=0.10)$  & -2.1779 & 0.02926 & $p<0.05^{*}$ \\
DE-CGAN $(\alpha=0.15)$  &  1.4438 & 0.1499 & $p>0.05$\\
DE-CGAN $(\alpha=0.20)$  & 0.4344 & 0.6672 & $p>0.05$  \\
CGAN $(\alpha=0.05)$     & 0.5443 & 0.5892 & $p>0.05$  \\
CGAN $(\alpha=0.10)$     & 0 & 1 & $p>0.05$ \\
CGAN $(\alpha=0.15)$     & 1.4438 & 0.1499 & $p>0.05$  \\
\bottomrule
\end{tabular}
\end{table}

\section{Discussion}\label{Discussion}
The current work shows the generation of synthetic psychiatric patients and their class labels improves the performance of a DNN classification model. These findings emphasize the importance of diverse datasets in deep learning and psychiatric research. This work introduces our novel framework DE-CGAN a novel method for the oversampling of difficult to classify fMRI connectivity features and synthetic class labels. These findings have significant implications in the field of psychiatry a field where the use of AI has been impacted by small sample sizes \cite{Koppe2020}.

Previous work has introduced methods for oversampling minority classes in classification tasks with imbalanced class labels. In the field of data mining this is referred to as the class imbalance problem \cite{Li2023}. The class imbalance problem occurs in classification tasks where examples for some classes are underrepresented when compared to the majority classes. In contrast, our problem deals with the under-representation of fMRI connectivity measures at certain bandwidths. Our work shows that identifying examples prone to misclassification and generating synthetic examples paired with a synthetic class label in the distribution of difficult to classify regions can lead to improvements in model performance. 

Artificial intelligence systems are being deployed rapidly in areas of non-trivial importance. Increasingly, researchers are exploring ways to deploy artificial intelligence systems to personalise medical care. Deep learning algorithms rely heavily on the data on which they are trained. However, problematically, the training data on which these algorithms rely is often not representative with insufficient examples of certain features. \citet{Tasci2022} assert under-representation of demographic, biological and outcome variables in data all contribute to algorithmic bias. 

Model performance is generally dependent on the data on which it is trained. As the artificial intelligence field pushes towards deployment of predictive models high-quality, representative data is essential \citet{Chen2021}. The importance of high quality data is twofold. Data that is truly representative of the general population is likely to improve the generalisation of models. An observed limitation of deep learning models. Additionally, data bias reduces the generalisability of artificial intelligence to unseen populations. Each of these issues is central to the fairness and equity of data science and its implementation in the field. 

The current work shows a training dataset of 10\% DE-CGAN examples achieves the optimal solution on a holdout test data set of real cases significantly more frequently than a model trained on real examples only. These works provide evidence for the generation of synthetic fMRI connectivity features and their class labels to over-sample datasets with difficult-to-classify patients due to under representation.

Furthermore, these findings emphasize that synthetically generated patients have the potential to supplement small datasets to improve model performance in small sample sizes. Additionally, these synthetic examples allow psychiatrists further data points to investigate relationships between variables and their impact on mental health.

While this work highlights the potential of synthetic records to increase the diversity of small datasets, this work also demonstrates as shows as the proportion of synthetic examples in the training data increases classification performance deteriorates. We propose two potential reasons for this performance decline: Firstly as the aim of DE-CGAN is to increase the proportion of underrepresented cases, as the proportion of these underrepresented cases makes up a greater proportion of the dataset the representation of the data population skews outside the true population such that the classification model is unable to classify the majority of examples. Secondly, it is possible that due to identified issues with GANs such as mode collapse, as the proportion of synthetic examples increases the synthetic data becomes less diverse. 

The general belief of generative networks is the synthetic data created comes from the same probability distribution as the training data from which they are created. However, as \citet{Jain2022} noted, that generated examples come from the same probability distribution is the best case. More likely, however, is generated examples are less diverse than the original data. As such, future work should explore explore further methods for generating diverse datasets for the purpose of sharing sufficiently diverse data between research organisations.

\section{Conclusion}\label{sec:conclusion}
This paper presents Diversity Enhancing Conditional Generative Adversarial Network, DE-CGAN, a novel framework for oversampling of underrepresented fMRI connectivity features and their class labels. Deep learning models require large and diverse datasets to perform optimally. When diverse datasets are not available model performance can start to deteriorate and is unlikely to generalise well to unseen data. In psychiatry, large datasets are difficult to source due to privacy and legal obligations. DE-GAN provides an option for balancing and extending datasets in psychiatry.

The current work demonstrates increasing the diversity of a training dataset of fMRI connectivity features with synthetic examples improves performance on a held-out test set of real examples. This work provides evidence for the viability of synthetically generated patients to increase the size and diversity of datasets which provides psychiatrists with more data to explore relationships between connectivity features and treatment outcomes. 

Future work should explore the potential of larger synthetic datasets for the study of rTMS response prediction. These synthetic datasets should maintain the characteristics of the original data to allow for sharing between research groups where legal obligations may prevent the sharing of actual data.


\bibliography{JIHR}

\end{document}